%% file: prob-embeddings.tex
\documentclass[11pt,letterpaper]{article}
\makeatletter
\newcommand{\@BIBLABEL}{\@emptybiblabel}
\newcommand{\@emptybiblabel}[1]{}
\makeatother
\usepackage[draft, hidelinks]{hyperref}
\usepackage{emnlp2016}
\usepackage{times}
\usepackage{amsmath}
\usepackage{amssymb}
\usepackage{graphicx}
\usepackage{latexsym}
\usepackage{booktabs}
\usepackage{color}
\usepackage{microtype}
\usepackage{mathtools}
\usepackage{mystuff-nlp}

\newcommand{\example}[1]{\textit{#1}}

\newcommand{\unk}{\ensuremath{\langle \textsc{unk} \rangle}}
\newcommand{\num}{\ensuremath{\langle \textsc{num} \rangle}}
\newcommand{\modelname}{\textsc{VarEmbed}}
\newcommand\qvec{\textsc{qvec}}
\newcommand\bbb{\textsc{SumEmbed}}
\newcommand\wtov{\textsc{word2vec}}

\DeclarePairedDelimiterX{\infdivx}[2]{(}{)}{%
  #1\;\delimsize\|\;#2%
}
\newcommand{\infdiv}{D_{KL}\infdivx}

\emnlpfinalcopy

\def\emnlppaperid{***}

\title{Morphological Priors for Probabilistic Neural Word Embeddings}

\author{Parminder Bhatia\thanks{\:\:The first two authors contributed equally. Code is available at \url{https://github.com/rguthrie3/MorphologicalPriorsForWordEmbeddings}.}\\
Yik Yak, Inc.\\
3525 Piedmont Rd NE, Building 6, Suite 500\\
Atlanta, GA\\
\texttt{parminder@yikyakapp.com}
\And
Robert Guthrie\footnotemark[1] \and Jacob Eisenstein\\
School of Interactive Computing\\
Georgia Institute of Technology\\
Atlanta, GA 30312 USA\\
\texttt{\{rguthrie3 + jacobe\}@gatech.edu}}

\date{}

\begin{document}
\maketitle

\begin{abstract}
\input{abstract}
\end{abstract}

\input{intro}
\input{model}

\input{implementation}
\input{evaluation}
\input{related}
\input{con}
\input{ack}

\citestr
\bibliographystyle{emnlp2016}

\end{document}

%% file: abstract.tex
Word embeddings allow natural language processing systems to share statistical information across related words. These embeddings are typically based on distributional statistics, making it difficult for them to generalize to rare or unseen words. We propose to improve word embeddings by incorporating morphological information, capturing shared sub-word features. Unlike previous work that constructs word embeddings directly from morphemes, we combine morphological and distributional information in a unified probabilistic framework, in which the word embedding is a latent variable. The morphological information provides a prior distribution on the latent word embeddings, which in turn condition a likelihood function over an observed corpus. This approach yields improvements on intrinsic word similarity evaluations, and also in the downstream task of part-of-speech tagging.

%% file: intro.tex
\section{Introduction}
Word embeddings have been shown to improve many natural language processing applications, from language models~\cite{mikolov2010recurrent} to information extraction~\cite{collobert2008unified}, and from parsing~\cite{chen2014fast} to machine translation~\cite{cho2014learning}. Word embeddings leverage a classical idea in natural language processing: use distributional statistics from large amounts of unlabeled data to learn representations that allow sharing across related words~\cite{brown1992class}. While this approach is undeniably effective, the long-tail nature of linguistic data ensures that there will always be words that are not observed in even the largest corpus~\cite{zipf1949human}. There will be many other words which are observed only a handful of times, making the distributional statistics too sparse to accurately estimate the 100- or 1000-dimensional dense vectors that are typically used for word embeddings. These problems are particularly acute in morphologically rich languages like German and Turkish, where each word may have dozens of possible inflections.

Recent work has proposed to address this issue by replacing word-level embeddings with embeddings based on subword units: morphemes~\cite{luong2013better,botha2014compositional} or individual characters~\cite{santos2014learning,ling2015finding,kim2016character}. Such models leverage the fact that word meaning is often \emph{compositional}, arising from subword components. By learning representations of subword units, it is possible to generalize to rare and unseen words.

But while morphology and orthography are sometimes a signal of semantics, there are also many cases similar spellings do \emph{not} imply similar meanings: \example{better}-\example{batter}, \example{melon}-\example{felon}, \example{dessert}-\example{desert}, etc. If each word's embedding is constrained to be a deterministic function of its characters, as in prior work, then it will be difficult to learn appropriately distinct embeddings for such pairs.
Automated morphological analysis may be incorrect: for example, \example{really} may be segmented into \example{re+ally}, incorrectly suggesting a similarity to \example{revise} and \example{review}. Even correct morphological segmentation may be misleading. Consider that \example{incredible} and \example{inflammable} share a prefix \example{in-}, which exerts the opposite effect in these two cases.\footnote{The confusion is resolved by morphologically analyzing the second example as \example{(in+flame)+able}, but this requires hierarchical morphological parsing, not just segmentation.}
Overall, a word's observed internal structure gives evidence about its meaning, but it must be possible to override this evidence when the distributional facts point in another direction.

We formalize this idea using the machinery of probabilistic graphical models. We treat word embeddings as \emph{latent variables}~\cite{vilnis2015word}, which are conditioned on a prior distribution that is based on word morphology. We then maximize a variational approximation to the expected likelihood of an observed corpus of text, fitting variational parameters over latent binary word embeddings. For common words, the expected word embeddings are largely determined by the expected corpus likelihood, and thus, by the distributional statistics. For rare words, the prior plays a larger role. Since the prior distribution is a function of the morphology, it is possible to impute embeddings for unseen words after training the model.

We model word embeddings as latent binary vectors. This choice is based on linguistic theories of lexical semantics and morphology.
Morphemes are viewed as adding morphosyntactic \emph{features} to words: for example, in English, \example{un-} adds a negation feature (\example{unbelievable}), \example{-s} adds a plural feature, and \example{-ed} adds a past tense feature~\cite{halle1993distributed}. 
Similarly, the lexicon is often viewed as organized in terms of features: for example, the word \example{bachelor} carries the features \textsc{human}, \textsc{male}, and \textsc{unmarried}~\cite{katz1963structure}. Each word's semantic role within a sentence can also be characterized in terms of binary features~\cite{dowty1991thematic,reisinger2015semantic}. Our approach is more amenable to such theoretical models than traditional distributed word embeddings. However, we can also work with the \emph{expected word embeddings}, which are vectors of probabilities, and can therefore be expected to hold the advantages of dense distributed representations~\cite{bengio2013representation}.

%% file: model.tex
\newcommand{\vb}[0]{\vec{b}}
\section{Model}
\label{sec:model}
\begin{figure*}
  \centering
  \includegraphics[width=.85\textwidth]{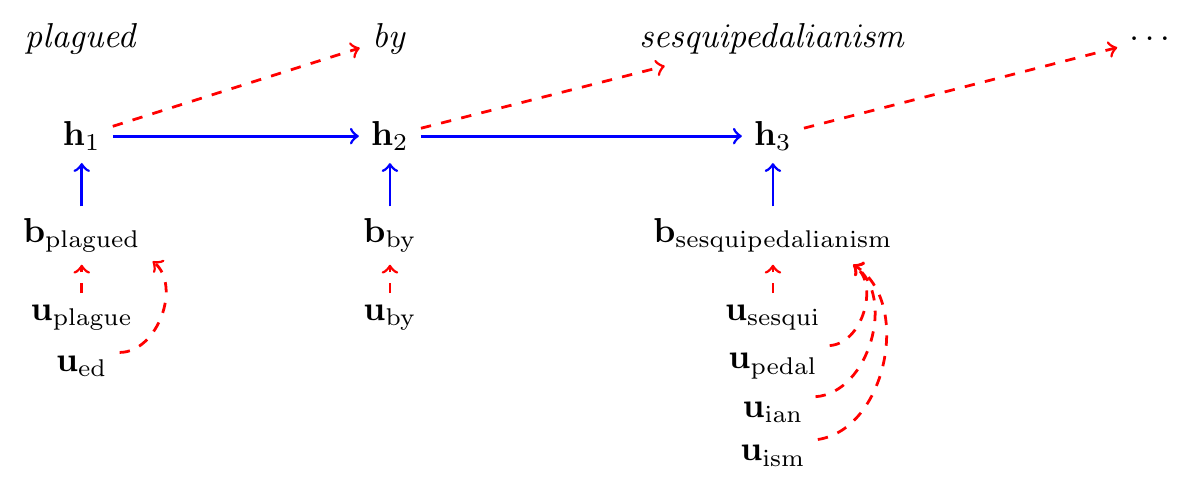}
  \caption{\textbf{Model architecture}, applied to the example sequence \example{\ldots plagued by sesquipedalianism \ldots}. Blue solid arrows indicate direct computation, red dashed arrows indicate probabilistic dependency. For simplicity, we present our models as recurrent neural networks rather than long short-term memories (LSTMs).}
  \label{fig:model-arch}
\end{figure*}

The modeling framework is illustrated in \autoref{fig:model-arch}, focusing on the word \example{sesquipedalianism}. This word is rare, but its morphology indicates several of its properties: the \example{-ism} suffix suggests that the word is a noun, likely describing some abstract property; the \example{sesqui-} prefix refers to one and a half, and so on. If the word is unknown, we must lean heavily on these intuitions, but if the word is well attested then we can rely instead on its examples in use. 

It is this reasoning that our modeling framework aims to formalize. We treat word embeddings as latent variables in a joint probabilistic model. The prior distribution over a word's embedding is conditioned on its morphological structure. The embedding itself then participates, as a latent variable, in a neural sequence model over a corpus, contributing to the overall corpus likelihood. If the word appears frequently, then the corpus likelihood dominates the prior --- which is equivalent to relying on the word's distributional properties. If the word appears rarely, then the prior distribution steps in, and gives a best guess as to the word's meaning.

Before describing these component pieces in detail, we first introduce some notation. The representation of word $w$ is a latent binary vector $\vb_w \in \{0,1\}^{k}$, where $k$ is the size of each word embedding. As noted in the introduction, this binary representation is motivated by feature-based theories of lexical semantics~\cite{katz1963structure}. Each word $w$ is constructed from a set of $M_w$ observed morphemes, $\mathcal{M}_w = (m_{w,1}, m_{w,2}, \ldots, m_{w,M_w})$. Each morpheme is in turn drawn from a finite vocabulary of size $v_m$, so that $m_{w,i} \in \{1, 2, \ldots, v_m\}$. Morphemes are obtained from an unsupervised morphological segmenter, which is treated as a black box. Finally, we are given a corpus, which is a sequence of words, $\vx = (x_1, x_2, \ldots, x_N)$, where each word $x_t \in \{1, 2, \ldots, v_w\}$, with $v_w$ equal to the size of the vocabulary, including the token \unk\ for unknown words.

\subsection{Prior distribution}
\label{sec:model-prior}
The key differentiating property of this model is that rather than estimating word embeddings directly, we treat them as a latent variable, with a prior distribution reflecting the word's morphological properties. To characterize this prior distribution, each morpheme $m$ is associated with an embedding of its own, $\vu_m \in \mathbb{R}^{k}$, where $k$ is again the embedding size. Then for position $i$ of the word embedding $\vb_w$, we have the following prior,
\begin{equation}
b_{w,i} \sim \text{Bernoulli}\left(\sigma(\sum_{m \in \set{M}_w} u_{m,i})\right),
\end{equation}
where $\sigma(\cdot)$ indicates the sigmoid function. The prior log-likelihood for a set of word embeddings is,
\begin{align}
\log & P(\vb; \set{M}, \vu) \\
= & \sum^{V_w}_w \log P(\vb_w ; \set{M}_w, \vu) \\
= & \sum^{V_w}_w \sum^k_i \log P(b_{w,i} ; \set{M}_w, \vu)\\
= & \sum^{V_w}_w \sum^k_i b_{w,i} \log \sigma \left( \sum_{m \in \set{M}_w} u_{m,i} \right)\\
\notag
& + (1 - b_{w,i}) \log \left(1 - \sigma \left( \sum_{m \in \set{M}_w} u_{m,i} \right)\right).
\end{align}

\subsection{Expected likelihood}
\label{sec:model-expected-ll}
The corpus likelihood is computed via a recurrent neural network language model~\cite[RNNLM]{mikolov2010recurrent}, which is a generative model of sequences of tokens. In the RNNLM, the probability of each word is conditioned on all preceding words through a recurrently updated state vector. This state vector in turn depends on the embeddings of the previous words, through the following update equations:
\begin{align}
\label{eq:hidden-update}
\vh_t = & f(\vb_{x_t}, \vh_{t-1}) \\
x_{t+1} \sim & \text{Multinomial}\left(\text{Softmax}\left[\mat{V} \vh_t\right]\right).
\end{align}

The function $f(\cdot)$ is a recurrent update equation; in the RNN, it corresponds to $\sigma(\Theta \vh_{t-1} + \vb_{x_t})$, where $\sigma(\cdot)$ is the elementwise sigmoid function. The matrix $\mat{V} \in \mathbb{R}^{v \times k}$ contains the ``output embeddings'' of each word in the vocabulary. We can then define the conditional log-likelihood of a corpus $\vx = (x_1, x_2, \ldots x_N)$ as,
\begin{align}
\log P(\vx \mid \vb) = & \sum^N_t \log P(x_t \mid \vh_{t-1}, \vb).
\end{align}

Since $\vh_{t-1}$ is deterministically computed from $\vx_{1:t-1}$ (conditioned on $\vb$), we can equivalently write the log-likelihood as,
\begin{equation}
\log P(\vx \mid \vb) =  \sum_t \log P(x_t \mid \vx_{1:t-1}, \vb).
\end{equation}

This same notation can be applied to compute the likelihood under a long-short term memory (LSTM) language model~\cite{sundermeyer2012lstm}. The only difference is that the recurrence function $f( \cdot )$ from \autoref{eq:hidden-update} now becomes more complex, including the input, output, and forget gates, and the recurrent state $\vh_t$ now includes the memory cell. As the LSTM update equations are well known, we focus on the more concise RNN notation, but we employ LSTMs in all experiments due to their better ability to capture long-range dependencies.

\subsection{Variational approximation}
Inference on the marginal likelihood $P(\vx_{1:N}) = \int P(\vx_{1:N}, \vb)  d\vb$ is intractable. We address this issue by making a variational approximation, 
\begin{align}
\log P(\vx) =& \log \sum_{\vb} P(\vx \mid \vb) P(\vb)\\
  = &  \log \sum_{\vb} \frac{Q(\vb)}{Q(\vb)} P(\vx \mid \vb) P(\vb)\\
  = &  \log E_q \left[P(\vx \mid \vb) \frac{P(\vb)}{Q(\vb)}\right]\\
  \geq & E_q [\log P(\vx \mid \vb) + \log P(\vb) - \log Q(\vb)]
\label{eq:var-bound}
\end{align}
The variational distribution $Q(\vb)$ is defined using a fully factorized mean field approximation,
\begin{align}
Q(\vb; \vec{\gamma}) = & \prod^{v_w}_w \prod^k_i q(b_{w,i} ; \gamma_{w,i}).
\end{align}
The variational distribution is a product of Bernoullis, with parameters $\gamma_{w,j} \in [0,1]$. In the evaluations that follow, we use the expected word embeddings $q(\vb_w)$, which are dense vectors in $[0,1]^{k}$. We can then use $Q(\cdot)$ to place a variational lower bound on the expected conditional likelihood,

Even with this variational approximation, the expected log-likelihood is still intractable to compute. In recurrent neural network language models, each word $x_t$ is conditioned on the entire prior history, $\vx_{1:t-1}$ --- indeed, this is one of the key advantages over fixed-length $n$-gram models. However, this means that the individual expected log probabilities involve not just the word embedding of $x_t$ and its immediate predecessor, but rather, the embeddings of \emph{all} words in the sequence $\vx_{1:t}$:
\begin{align}
 E_q & \left[\log P(\vx \mid \vb)\right] \\
 =& \sum^N_t \eq{\log P(x_t \mid \vx_{1:t-1}, \vb)} \\
\notag
=& \sum^N_t \sum_{\{\vb_w : w \in \vx_{1:t}\}} Q(\{\vb_w : w \in \vx_{1:t}\})\\ 
& \times \log P(x_t \mid \vx_{1:t-1}, \vb).
\end{align}

We therefore make a further approximation by taking a local expectation over the recurrent state,
\begin{align}
\eq{\vh_t} & \approx  f(\eq{\vb_{x_t}},\eq{\vh_{t-1}}) \\
 \eq{\log P(x_t \mid \vx_{1:t-1}, \vb)} & \approx \log \text{Softmax}\left(\mat{V} \eq{\vh_t}\right).
\label{eq:approx-ll}
\end{align}

This approximation means that we do not propagate uncertainty about $\vh_t$ through the recurrent update or through the likelihood function, but rather, we use local point estimates. Alternative methods such as variational autoencoders~\cite{chung2015recurrent} or sequential Monte Carlo~\cite{de2000sequential} might provide better and more principled approximations, but this direction is left for future work.

Variational bounds in the form of \autoref{eq:var-bound} can generally be expressed as a difference between an expected log-likelihood term and a term for the Kullback-Leibler (KL) divergence between the prior distribution $P(\vb)$ and the variational distribution $Q(\vb)$~\cite{wainwright2008graphical}. Incorporating the approximation in \autoref{eq:approx-ll}, the resulting objective is,
\begin{align}
\notag
\mathcal{L} = & \sum^N_t \log P(x_t \mid \vx_{1:{t-1}} ; E_q [\vb]) - \infdiv{Q(\vb)}{P(\vb)}.
\label{eq:bound-kl}
\end{align}

The KL divergence is equal to,
\begin{align}
& \infdiv{Q(\vb)}{P(\vb)}\\
\quad & =  \sum^{v_w}_w \sum^k_i \infdiv{q(b_{w,i})}{P(b_{w,i})} \\
\notag
 & = \sum^{v_w}_w \sum^k_i \gamma_{w,i} \log \sigma ( \sum_{m \in \set{M}_w} u_{m,i}) \\
\notag
 & \phantom{=} + (1 -  \gamma_{w,i}) \log (1 - \sigma( \sum_{m \in \set{M}_w} u_{m,i} ))\\
& \phantom{=} - \gamma_{w,i} \log \gamma_{w,i} - (1-\gamma_{w,i}) \log (1-\gamma_{w,i}).
\end{align}

Each term in the variational bound can be easily constructed in a computation graph, enabling automatic differentiation and the application of standard stochastic optimization techniques.

%% file: implementation.tex
\section{Implementation}
\label{sec:implementation}
The objective function is given by the variational lower bound in \autoref{eq:bound-kl}, using the approximation to the conditional likelihood described in \autoref{eq:approx-ll}. This function is optimized in terms of several parameters:
\begin{itemize}
\item the morpheme embeddings, $\{\vu_m\}_{m \in 1\ldots v_m}$;
\item the variational parameters on the word embeddings, $\{\vec{\gamma}\}_{w \in 1 \ldots v_w}$;
\item the output word embeddings $\mat{V}$;
\item the parameter of the recurrence function, $\Theta$.
\end{itemize}
Each of these parameters is updated via the \texttt{RMSProp} online learning algorithm~\cite{tieleman2012rmsprop}. The model and baseline (described below) are implemented in \texttt{blocks}~\cite{blocks2015}. In the remainder of the paper, we refer to our model as \modelname.

\subsection{Data and preprocessing}
All embeddings are trained on 22 million tokens from the the North American News Text (NANT) corpus~\cite{graff1995north}. We use an initial vocabulary of 50,000 words, with a special \unk\  token for words that are not among the 50,000 most common. We then perform downcasing and convert all numeric tokens to a special \num\ token. After these steps, the vocabulary size decreases to 48,986. Note that the method can impute word embeddings for out-of-vocabulary words under the prior distribution $P(\vb ; \set{M}, \vu)$; however, it is still necessary to decide on a vocabulary size to determine the number of variational parameters $\vec{\gamma}$ and output embeddings to estimate.

Unsupervised morphological segmentation is performed using Morfessor~\cite{creutz2002unsupervised}, with a maximum of sixteen morphemes per word. This results in a total of 14,000 morphemes, which includes stems for monomorphemic words. We do not rely on any labeled information about morphological structure, although the incorporation of gold morphological analysis is a promising topic for future work.

\subsection{Learning details} The LSTM parameters are initialized uniformly in the range $[-0.08, 0.08]$. The word embeddings are initialized using pre-trained \texttt{word2vec} embeddings. We train the model for 15 epochs, with an initial learning rate of $0.01$, a decay of $0.97$ per epoch, and minibatches of size $25$. We clip the norm of the gradients (normalized by minibatch size) at 1, using the default settings in the \texttt{RMSprop} implementation in \texttt{blocks}. These choices are motivated by prior work~\cite{zaremba2014recurrent}. After each iteration, we compute the objective function on the development set; when the objective does not improve beyond a small threshold, we halve the learning rate. 

Training takes roughly one hour per iteration using an NVIDIA 670 GTX, which is a commodity graphics processing unit (GPU) for gaming. This is nearly identical to the training time required for our reimplementation of the algorithm of \newcite{botha2014compositional}, described below.

\subsection{Baseline}
\label{sec:baseline}
The most comparable approach is that of \newcite{botha2014compositional}. In their work, embeddings are estimated for each morpheme, as well as for each in-vocabulary word. The final embedding for a word is then the sum of these embeddings, e.g., 
\begin{equation}
\overline{\strut \text{greenhouse}} = \overline{\strut \textit{greenhouse}} + \overline{\strut \textit{green}} + \overline{\strut \textit{house}},
\end{equation}
where the italicized elements represent learned embeddings. 

We build a baseline that is closely inspired by this approach, which we call \bbb. The key difference is that while \newcite{botha2014compositional} build on the log-bilinear language model~\cite{mnih2007three}, we use the same LSTM-based architecture as in our own model implementation. This enables our evaluation to focus on the critical difference between the two approaches: the use of latent variables rather than summation to model the word embeddings. As with our method, we used pre-trained \texttt{word2vec} embeddings to initialize the model.

\subsection{Number of parameters} The dominant terms in the overall number of parameters are the (expected) word embeddings themselves. The variational parameters of the input word embeddings, $\gamma$, are of size $k \times v_w$. The output word embeddings are of size $\#|\vh| \times v_w$. The morpheme embeddings are of size $k \times v_m$, with $v_m \ll v_w$. In our main experiments, we set $v_w = 48,896$ (see above), $k = 128$, and $\#|\vh| = 128$. After including the character embeddings and the parameters of the recurrent models, the total number of parameters is roughly $12.8$ million. This is identical to number of parameters in the \bbb\ baseline. 

%% file: evaluation.tex
\begin{table*}
  \centering
  \begin{tabular}{llll}
    \toprule
    & \wtov & \bbb & \modelname \\
    \midrule
    Wordsim353 \\
    all words (353) & n/a & 42.9 & \textbf{48.8} \\
    in-vocab (348) & \textbf{51.4} & 45.9 & 51.3 \\[3pt]
    rare words (rw) \\
    all words (2034) & n/a & 23.0 & \textbf{24.0} \\
    in-vocab (715) & 33.6 & 37.3 & \textbf{44.1}  \\
    \bottomrule
  \end{tabular}
  \caption{Word similarity evaluation results, as measured by Spearmann's $\rho \times 100$. \wtov\  cannot be evaluated on all words, because embeddings are not available for out-of-vocabulary words. The total number of words in each dataset is indicated in parentheses.}
  \label{tab:wordsim}
\end{table*}

\section{Evaluation}
\label{sec:evaluation}
Our evaluation compares the following embeddings:
\begin{description}
\item[\wtov] We train the popular \texttt{word2vec} CBOW (continuous bag of words) model~\cite{mikolov2013distributed}, using the \texttt{gensim} implementation.
\item[\bbb] We compare against the baseline described in \autoref{sec:baseline}, which can be viewed as a reimplementation of the compositional model of \newcite{botha2014compositional}.
\item[\modelname] For our model, we take the \emph{expected} embeddings $E_q[\vb]$, and then pass them through an inverse sigmoid function to obtain values over the entire real line.
\end{description}


\subsection{Word similarity}
Our first evaluation is based on two classical word similarity datasets: Wordsim353~\cite{finkelstein2001placing} and the Stanford ``rare words'' (rw) dataset~\cite{luong2013better}. We report Spearmann's $\rho$, a measure of rank correlation, evaluating on both the entire vocabulary as well as the subset of in-vocabulary words.

As shown in \autoref{tab:wordsim}, \modelname\ consistently outperforms \bbb\ on both datasets. On the subset of in-vocabulary words, \wtov\ gives slightly better results on the wordsim words that are in the NANT vocabulary, but is not applicable to the complete dataset. On the rare words dataset, \wtov\ performs considerably worse than both morphology-based models, matching the findings of \newcite{luong2013better} and \newcite{botha2014compositional} regarding the importance of morphology for doing well on this dataset.

\subsection{Alignment with lexical semantic features}
Recent work questions whether these word similarity metrics are predictive of performance on downstream tasks~\cite{faruqui2016problems}. The \qvec\ statistic is another intrinsic evaluation method, which has been shown to be better correlated with downstream tasks~\cite{tsvetkov2015evaluation}. This metric measures the alignment between word embeddings and a set of lexical semantic features. Specifically, we use the semcor noun verb supersenses oracle provided at the qvec github repository.\footnote{\url{https://github.com/ytsvetko/qvec}}

\begin{table}
  \centering
  \begin{tabular}{lp{0.7in}p{0.7in}}
    \toprule
    & all words (4199) & in vocab (3997) \\
    \midrule
    \wtov & n/a & \textbf{34.8} \\
    \bbb & 32.8 & 33.5 \\
    \modelname & \textbf{33.6} & 34.7 \\[3pt]
    morphemes only\\
    \bbb & 24.7 & 25.1\\
    \modelname & \textbf{30.2} & \textbf{31.0} \\
    \bottomrule
  \end{tabular}
  \caption{Alignment with lexical semantic features, as measured by \qvec. Higher scores are better, with a maximum possible score of $100$.}
  \label{tab:qvec}
\end{table}

As shown in \autoref{tab:qvec}, \modelname\ outperforms \bbb\ on the full lexicon, and gives similar performance to \wtov\ on the set of in-vocabulary words. 
We also consider the morpheme embeddings alone. For \bbb, this means that we construct the word embedding from the sum of the embeddings for its morphemes, \emph{without} the additional embedding per word. For \modelname, we use the expected embedding under the prior distribution $E[\vb \mid \vc]$. The results for these representations are shown in the bottom half of \autoref{tab:qvec}, revealing that \modelname\  learns much more meaningful embeddings at the morpheme level, while much of the power of \bbb\  seems to come from the word embeddings.


\subsection{Part-of-speech tagging}
Our final evaluation is on the downstream task of part-of-speech tagging, using the Penn Treebank. We build a simple classification-based tagger, using a feedforward neural network. (This is not intended as an alternative to state-of-the-art tagging algorithms, but as a comparison of the syntactic utility of the information encoded in the word embeddings.) The inputs to the network are the concatenated embeddings of the five word neighborhood $(x_{t-2},x_{t-1},x_t,x_{t+1},x_{t+2})$; as in all evaluations, 128-dimensional embeddings are used, so the total size of the input is 640. This input is fed into a network with two hidden layers of size $625$, and a softmax output layer over all tags. We train using RMSProp~\cite{tieleman2012rmsprop}. 

\begin{figure}
  \centering
  \includegraphics[width=0.57\textwidth]{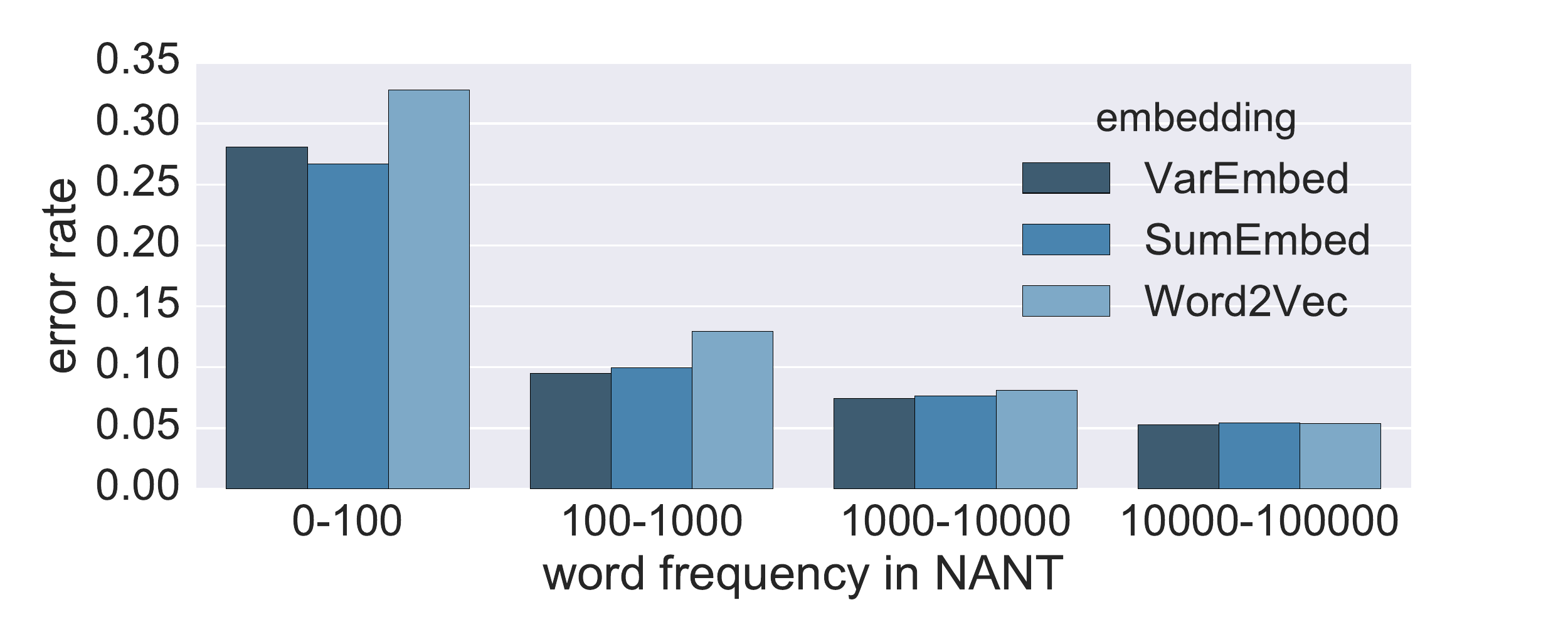}
  \caption{Error rates by word frequency.}
  \label{fig:pos-err}
\end{figure}

Results are shown in \autoref{tab:pos}. Both morphologically-informed embeddings are significantly better to \wtov\ ($p < .01$, two-tailed binomial test), but the difference between \bbb\ and \modelname\  is not significant at $p < .05$. \autoref{fig:pos-err} breaks down the errors by word frequency. As shown in the figure, the tagger based on \wtov\ performs poorly for rare words, which is expected because these embeddings are estimated from sparse distributional statistics. \bbb\ is slightly better on the rarest words (the $0-100$ group accounts for roughly 10\% of all tokens). In this case, it appears that this simple additive model is better, since the distributional statistics are too sparse to offer much improvement. The probabilistic \modelname\ embeddings are best for all other frequency groups, showing that it effectively combines morphology and distributional statistics.

\begin{table}
  \centering
  \begin{tabular}{lll}
    \toprule
    & dev & test  \\
    \midrule
    \wtov & 92.42 & 92.40 \\
    \bbb & \textbf{93.26} & \textbf{93.26}\\
    \modelname & 93.05 & 93.09\\
    \bottomrule
  \end{tabular}
  \caption{Part-of-speech tagging accuracies}
  \label{tab:pos}
\end{table}

%% file: related.tex
\section{Related work}
\label{sec:related}
\paragraph{Adding side information to word embeddings} An alternative approach to incorporating additional information into word embeddings is to constrain the embeddings of semantically-related words to be similar. Such work typically draws on existing lexical semantic resources such as WordNet. For example, \newcite{yu2014improving} define a joint training objective, in which the word embedding must predict not only neighboring word tokens in a corpus, but also related word \emph{types} in a semantic resource; a similar approach is taken by \newcite{bian2014knowledge}. Alternatively, \newcite{faruqui2015retrofitting} propose to ``retrofit'' pre-trained word embeddings over a semantic network. Both retrofitting and our own approach treat the true word embeddings as latent variables, from which the pre-trained word embeddings are stochastically emitted. However, a key difference from our approach is that the underlying representation in these prior works is relational, and not generative. These methods can capture similarity between words in a relational lexicon such as WordNet, but they do not offer a generative account of how (approximate) meaning is constructed from orthography or morphology. 

\paragraph{Word embeddings and morphology} The \bbb\ baseline is based on the work of \newcite{botha2014compositional}, in which words are segmented into morphemes using \textsc{Morfessor}~\cite{creutz2002unsupervised}, and then word representations are computed through addition of morpheme representations. A key modeling difference from this prior work is that rather than computing word embeddings directly and deterministically from subcomponent embeddings (morphemes or characters, as in~\cite{ling2015finding,kim2016character}), we use these subcomponents to define a prior distribution, which can be overridden by distributional statistics for common words. Other work exploits morphology by training word embeddings to optimize a joint objective over distributional statistics and rich, morphologically-augmented part of speech tags~\cite{cotterell2015morphological}. This can yield better word embeddings, but does not provide a way to compute embeddings for unseen words, as our approach does. 

Recent work by \newcite{cotterell2016morphological} extends the idea of retrofitting, which was based on semantic similarity, to a morphological framework. In this model, embeddings are learned for morphemes as well as for words, and each word embedding is conditioned on the sum of the morpheme embeddings, using a multivariate Gaussian. The covariance of this Gaussian prior is set to the inverse of the number of examples in the training corpus, which has the effect of letting the morphology play a larger role for rare or unseen words. Like all retrofitting approaches, this method is applied in a pipeline fashion after training word embeddings on a large corpus; in contrast, our approach is a joint model over the morphology and corpus. Another practical difference is that \newcite{cotterell2016morphological} use gold morphological features, while we use an automated morphological segmentation. 

\paragraph{Latent word embeddings} Word embeddings are typically treated as a parameter, and are optimized through point estimation~\cite{bengio2003neural,collobert2008unified,mikolov2010recurrent}. Current models use word embeddings with hundreds or even thousands of parameters per word, yet many words are observed only a handful of times. It is therefore natural to consider whether it might be beneficial to model uncertainty over word embeddings. \newcite{vilnis2015word} propose to model Gaussian densities over dense vector word embeddings. They estimate the parameters of the Gaussian directly, and, unlike our work, do not consider using orthographic information as a prior distribution. This is easy to do in the latent binary framework proposed here, which is also a better fit for some theoretical models of lexical semantics~\cite{katz1963structure,reisinger2015semantic}. This view is shared by \newcite{kruszewski2015deriving}, who induce binary word representations using labeled data of lexical semantic entailment relations, and by \newcite{henderson2016vector}, who take a mean field approximation over binary representations of lexical semantic features to induce hyponymy relations.

More broadly, our work is inspired by recent efforts to combine directed graphical models with discriminatively trained ``deep learning'' architectures. The variational autoencoder~\cite{kingma2013auto}, neural variational inference~\cite{mnih2014neural,miao2016neural}, and black box variational inference~\cite{ranganath2014black} all propose to use a neural network to compute the variational approximation. These ideas are employed by \newcite{chung2015recurrent} in the variational recurrent neural network, which places a latent continuous variable at each time step. In contrast, we have a dictionary of latent variables --- the word embeddings --- which introduce uncertainty over the hidden state $\vh_t$ in a standard recurrent neural network or LSTM. We train this model by employing a mean field approximation, but these more recent techniques for neural variational inference may also be applicable. We plan to explore this possibility in future work.

%% file: con.tex
\section{Conclusion and future work}
We present a model that unifies compositional and distributional perspectives on lexical semantics, through the machinery of Bayesian latent variable models. In this framework, our prior expectations of word meaning are based on internal structure, but these expectations can be overridden by distributional statistics. The model is based on the very successful long-short term memory (LSTM) for sequence modeling, and while it employs a Bayesian justification, its inference and estimation are little more complicated than a standard LSTM. This demonstrates the advantages of reasoning about uncertainty even when working in a ``neural'' paradigm.

This work represents a first step, and we see many possibilities for improving performance by extending it. Clearly we would expect this model to be more effective in languages with richer morphological structure than English, and we plan to explore this possibility in future work. From a modeling perspective, our prior distribution merely sums the morpheme embeddings, but a more accurate model might account for sequential or combinatorial structure, through a recurrent~\cite{ling2015finding}, recursive~\cite{luong2013better}, or convolutional architecture~\cite{kim2016character}. There appears to be no technical obstacle to imposing such structure in the prior distribution. Furthermore, while we build the prior distribution from morphemes, it is natural to ask whether characters might be a better underlying representation: character-based models may generalize well to non-word tokens such as names and abbreviations, they do not require morphological segmentation, and they require a much smaller number of underlying embeddings. On the other hand, morphemes encode rich regularities across words, which may make a morphologically-informed prior easier to learn than a prior which works directly at the character level. It is possible that this tradeoff could be transcended by combining characters and morphemes in a single model. 

Another advantage of latent variable models is that they admit partial supervision. If we follow \newcite{tsvetkov2015evaluation} in the argument that word embeddings should correspond to lexical semantic features, then an inventory of such features could be used as a source of partial supervision, thus locking dimensions of the word embeddings to specific semantic properties. This would complement the graph-based ``retrofitting'' supervision proposed by \newcite{faruqui2015retrofitting}, by instead placing supervision at the level of individual words.

%% file: ack.tex
\section*{Acknowledgments}
Thanks to Erica Briscoe, Martin Hyatt, Yangfeng Ji, Bryan Leslie Lee, and Yi Yang for helpful discussion of this work. Thanks also the EMNLP reviewers for constructive feedback. This research is supported by the Defense Threat Reduction Agency under award HDTRA1-15-1-0019.